\documentclass[11pt]{article}

\usepackage[final]{acl}

\usepackage{times}
\usepackage{latexsym}

\usepackage[T1]{fontenc}

\usepackage[utf8]{inputenc}

\usepackage{microtype}

\usepackage{inconsolata}

\usepackage{graphicx}

\title{Pair Difficulty Matters: Rethinking Pairwise LLM-as-a-Judge\\Evaluation and Consistency}

\author{Bruno Brocai \\
  Heidelberg University \\
  German Department \\
  \texttt{bruno.brocai@gs.uni-heidelberg.de      } \\\And
  Maria Becker \\
  Heidelberg University \\
  German Department \\
  \texttt{     maria.becker@gs.uni-heidelberg.de}
}

\usepackage{booktabs}
\usepackage{makecell}
\usepackage{tcolorbox}
\tcbuselibrary{breakable}
\usepackage{multirow}
\usepackage{amsmath}

\newtcolorbox{promptbox}[1]{
  colback=gray!5, colframe=gray!60,
  title=#1, fonttitle=\bfseries\small,
  breakable, boxrule=0.5pt, arc=1mm,
  left=1mm, right=1mm, fontupper=\small\ttfamily
}

\begin{document}
\maketitle
\begin{abstract}
Large Language Model judges are widely used to rank texts and
text-generating systems through pairwise comparison, and their reliability
is typically assessed via three proxies: position bias, transitivity, and
pairwise agreement (self- or human-labeled). Because these proxies drive
judge selection and benchmarking, a substantial literature reporting that
judges perform poorly on them risks steering practitioners away from
otherwise capable evaluators. We argue this assessment is misleading.
Under the Bradley--Terry geometry underlying pairwise aggregation, each proxy is dominated by close-rank-gap pairs, where inconsistency is expected and  errors cost little aggregate ranking accuracy; far-gap pairs determine the ranking accuracy but barely move the proxies. We formalize this argument and validate it in
a controlled simulation and on two human-rated corpora: the proxies
correlate only weakly with ranking accuracy against gold, and their
predictive component concentrates in the far-gap regime. Judges should
therefore be assessed on rank-gap-conditional metrics, ideally against
human rankings. Code at \href{https://github.com/brunobrocai/PairDifficulty}{https://github.com/brunobrocai/PairDifficulty}.
\end{abstract}

\section{Introduction}
LLM-as-a-judge has become a central paradigm for evaluating generated text across many domains \cite{guSurveyLLMasaJudge2025a}. However, a growing body of work has documented systematic problems with LLM-judges that fall broadly under the heading of inconsistency, and several distinct forms have been identified. Models exhibit run-to-run inconsistency, returning different verdicts across repeated queries under identical or near-identical settings. They exhibit position bias --- by far the most-discussed failure mode --- preferring one slot of a pairwise comparison at rates well above chance. They also exhibit non-transitivity at the level of the induced preference graph.

A separate line of work has compared LLM verdicts not to internal consistency criteria but to individual pairwise judgments, focusing on human–LLM divergence at the pair level \cite[e.g.,][]{feinLitBenchBenchmarkDataset2026, zhengJudgingLLMasaJudgeMTBench2023a, liGenerativeJudgeEvaluating2023}. These two strands --- internal inconsistency on the one hand, and disagreement with humans on the other --- are typically treated as distinct problems. However, the comparisons that dominate aggregate inconsistency metrics are often the comparisons that matter least for ranking recovery. We argue that they are closely connected: They are both affected by pair gap / difficulty, and because of this, are both unreliable proxies of judge quality when it comes to recovering human-derived rankings.

\section{Related Work}

Swapping two responses can flip a judge's
verdict --- a phenomenon documented early on by
\citet{wangLargeLanguageModels2024} and
\citet{zhengJudgingLLMasaJudgeMTBench2023a}, and usually addressed by
balanced-position aggregation.
\citet{shiJudgingJudgesSystematic2025} provide the most thorough
analysis to date and find that the gap between texts is a key factor:
pairs close in quality are affected far more strongly than distant ones.

\citet{xuInvestigatingNonTransitivityLLMasaJudge2025} document
non-transitivity in pairwise rankings and show that, like position
bias, it concentrates on close pairs; they mitigate it through round-robin
tournaments and Bradley--Terry (BT) aggregation.
\citet{wangTrustJudgeInconsistenciesLLMasaJudge2025} target
inconsistency directly and document that 85–90\% of transitivity violations are tie-driven and that judge capability does not monotonically reduce inconsistency. SAGE
\cite{fengSageScalableFramework2025} evaluates judges along exactly these two axes without requiring human-labeled comparisons, and finds
that even top models fail on roughly a quarter of difficult cases.
All of this work treats inconsistency as a defect that needs to be reduced rather
than asking what its structure reveals about the judge.

\citet{zhengJudgingLLMasaJudgeMTBench2023a} measure pair-level
agreement between LLM judges and humans directly and find that this agreement is
gap-conditional: the residual disagreement
concentrates on close pairs where humans also disagree.
\citet{thakurJudgingJudgesEvaluating2025} argue that aggregate
alignment metrics are misleading proxies for judge usefulness: judges
with substantially lower percent agreement can still produce ranking
correlations with humans that are nearly identical, because consistent
biases preserve relative ordering even when they distort absolute
scores. We extend this line of argument along an orthogonal axis.
While \citeauthor{thakurJudgingJudgesEvaluating2025} decompose by use case (scoring versus ranking),
we decompose by the rank gap of the items being compared, and show
that aggregate inconsistency is dominated by close-pair behavior that
a calibrated judge often produces.

\section{Theory}
\label{sec:theory}

We aggregate pairwise verdicts under the BT model \cite{bradleyRankAnalysisIncomplete1952}, the most robust ranking algorithm for most use cases \cite{daynauthRankingUnraveledRecipes2025}. Under BT, each item $i$ carries a latent strength $\theta_i$, and the probability that $i$ is preferred to $j$ is $\Pr(i \succ j) = \sigma(\theta_i - \theta_j)$, where $\sigma(x) = (1 + e^{-x})^{-1}$ is the logistic function. In a pair with a true strength gap $\Delta = \theta_i - \theta_j$, the two possible outcomes have log-probabilities $\log\sigma(\Delta)$ and $\log\sigma(-\Delta)$, which differ by exactly $|\Delta|$. Close pairs are near-coinflips, and the two outcomes are nearly equally consistent with the model; far pairs are lopsided, and one outcome is much more consistent than the other. A flipped verdict therefore changes that pair's log-likelihood contribution by $|\Delta|$:
\begin{itemize}
  \item \textbf{Close pair} ($\Delta \approx 0$): both outcomes have nearly the same likelihood, so a flip is nearly free under the model.
  \item \textbf{Far pair} (large $|\Delta|$): the two outcomes differ in log-likelihood by $|\Delta|$, so a flip is strong evidence of judge error.
\end{itemize}
This separates two failure modes that aggregate metrics conflate. Close-pair \emph{self-inconsistency} --- the judge flipping its own verdict across runs --- is expected noise: the BT model itself assigns the two outcomes near-equal probability, so a flip is consistent with the model, and averages out across comparisons. Far-pair self-inconsistency, by contrast, is where calibration error becomes visible: the model strongly predicts one outcome, so a flip is a real error, and such errors tend to push an item’s score consistently in one direction. Noise of the first kind does not systematically bias the recovered ranking; error of the second kind compresses the recovered scale and yields a worse ranking at matched comparison budget. The same $|\Delta|$-weighting applies to judge--human \emph{disagreement}: when the judge’s verdict diverges from the human verdict, the disagreement is again most diagnostic in the high-$|\Delta|$ regime, while close-pair disagreement is bounded below by the noise human annotators themselves cannot avoid, so under a BT-behaving human reference, close-pair observations carry little information about judge quality. Pair-level agreement metrics --- Krippendorff's $\alpha$, raw agreement rates, position-flip rates --- weight every pair equally and therefore systematically overweight the regime that contributes least to ranking distance from gold, whether the comparison is judge-vs-judge or judge-vs-human.

This predicts the empirical pattern: pair-level reliability and
position-bias scores should correlate weakly with ranking accuracy
against human gold, with predictive signal concentrated in the
far-pair tail. It also clarifies prior findings: position bias is
empirically strongest on close pairs
\citep{shiJudgingJudgesSystematic2025}, exactly where it costs the ranking least, and human annotators themselves disagree
most on close pairs \citep{zhengJudgingLLMasaJudgeMTBench2023a} ---
as BT predicts. Judges and humans agree on what the hard cases are;
the question this paper takes up is whether the measures recognize that the hard cases
are also the cheap ones.

\section{Simulation: Human Agreement}
\label{sim}

To isolate the effect of disagreement distribution independently of model-specific behavior, we construct a controlled simulation in which judges differ only in where across the rank-gap spectrum their errors occur.

We simulate a dataset with normally distributed BT strengths and generate human comparisons from randomly drawing the winner of each comparison by using BT-probability (see appendix~\ref{app:sim} for results with other simulation settings). Then we simulate models based on the human comparisons to ensure exactly equal pairwise agreement. The simulated models pick the human's winner 60\% of the time --- they only differ on which pair gold gaps these errors concentrate. We do that by sampling error pairs with different weights, where weight $ q = 0$ means the disagreement is completely evenly distributed and $ q = 1 $ means all disagreement is concentrated on far pairs.

\begin{figure}[t]
  \centering
  \includegraphics[width=1\linewidth]{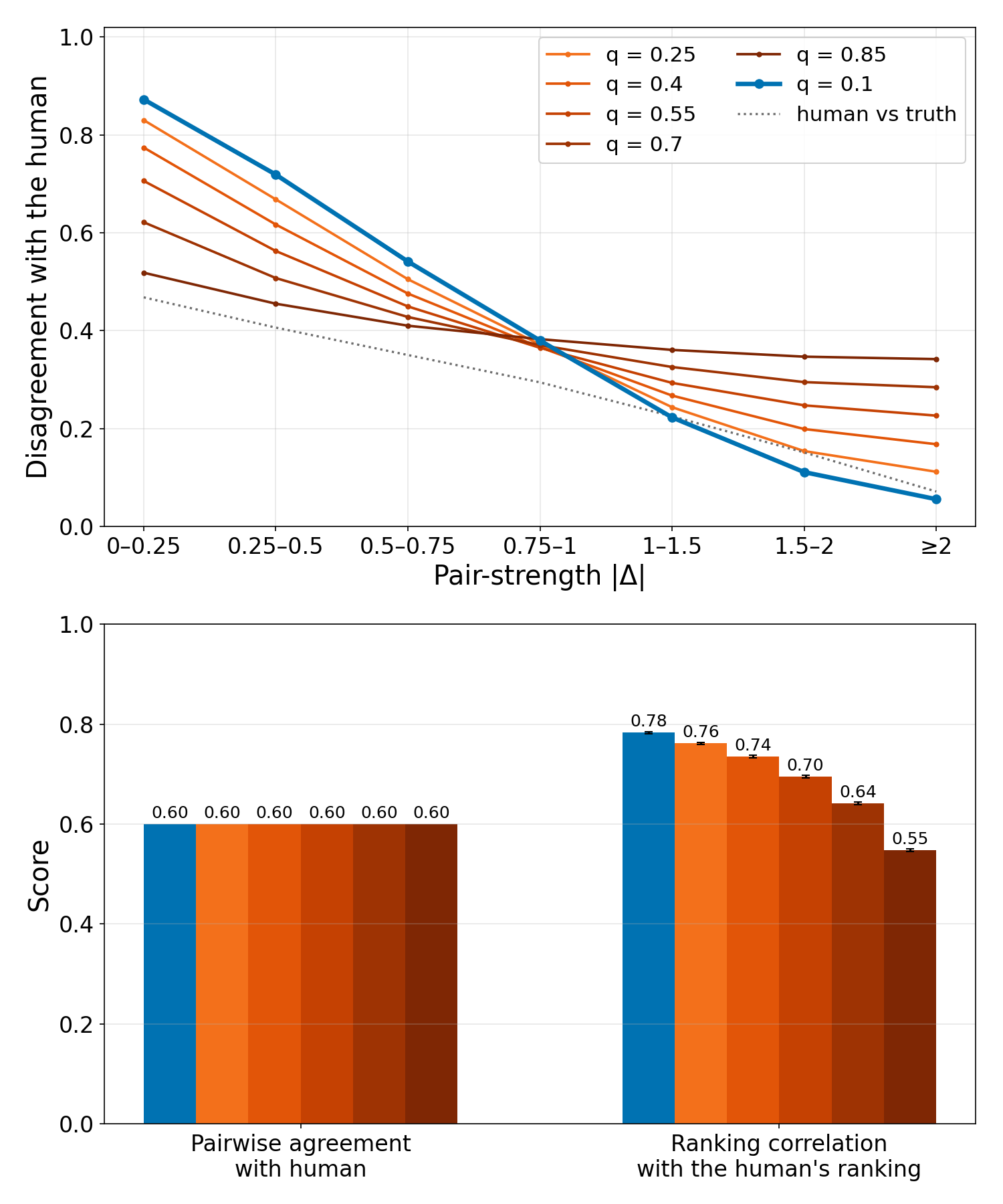}
  \caption{Simulation results for BT Judges with identical pairwise agreement. Top shows how disagreement between judges and humans is differently distributed, while overall agreement is exactly equal (bottom left). Despite this, ranking correlation with the human gold varies substantially depending on where disagreement occurs (bottom right).}
  \label{fig:Simulation Results}
\end{figure}

As Figure~\ref{fig:Simulation Results} shows, pairwise agreement between simulated judges and humans is equal across different $q$ --- the only difference is that some judges agree less with humans on distant pairs, while others more on close pairs. Despite that, agreement with gold falls sharply as judge errors increase at higher $|\Delta|$s, even if these errors are counterbalanced by higher accuracy on close pairs. Therefore, pairwise-level agreement is not the same as ranking correlation due to the spread of disagreement, and cannot substitute for ranking correlation. Namely, it cannot distinguish a judge that performs well on easy/far pairs but uses a misaligned heuristic (e.g. length) to tie-break close pairs from a judge doing exactly the opposite, but the former is a much better judge.

\section{Experiment: Model Inconsistencies}

\paragraph{Datasets} \textsc{Clear} \cite{crossleyCommonLitEaseReadability} is a dataset of English school text excerpts rated on comprehensibility by teachers. The human rating was performed pairwise: teachers were shown two texts and asked to pick which one is easier for students to understand. We use 200 randomly selected texts. \textsc{ASAP 2.0} \cite{crossleyLargescaleCorpusAssessing2025a} is a corpus of student writing tests for grades 6 to 10, where students were tasked to argue for a position on a topic using a source text on that topic. The essays are human-rated on a scale of 1--6. We use 200 student essays from the test set on the topic of algorithmic facial emotion detection for grade 10.

Ranking human-labeled corpus texts rather than model outputs provides a clean gold standard for pairwise rank gaps, which is unavailable when the ranked items are the generating models themselves.

\begin{table*}
  \centering
  \begin{tabular}{llcccccccc}
    \toprule
                                                     &                 & \multicolumn{3}{c}{Consistency ($1-$flip)} & IPP      & \multicolumn{3}{c}{\% non-transitive triads} &                                    \\
    \cmidrule(lr){3-5}\cmidrule(lr){7-9}
                                                     & Judge           & all                                        & close    & far                                          & all  & cyc & mix & ineq & $\rho$   \\
                                                     &                 &                                            & (Q1 gap) & (Q4 gap)                                     &      &     &     &      &          \\
    \midrule
    \multirow{6}{*}{\rotatebox[origin=c]{90}{CLEAR}} & GPT-5.4-nano    & .78                                        & .72      & .90                                          & .61  & 0.0 & 0.8 & 9.9  & $+0.691$ \\
                                                     & GPT-5.4-mini    & .80                                        & .66      & .96                                          & .97  & 0.0 & 0.4 & 9.5  & $+0.825$ \\
                                                     & Ministral-3-3B  & .79                                        & .70      & .92                                          & .89  & 0.0 & 0.4 & 6.2  & $+0.755$ \\
                                                     & Ministral-3-14B & .84                                        & .73      & .97                                          & .89  & 0.0 & 0.8 & 4.1  & $+0.788$ \\
                                                     & Gemma-3-4B      & .22                                        & .09      & .46                                          & 1.00 & 0.0 & 0.0 & 27.2 & $+0.676$ \\
                                                     & Gemma-3-12B     & .76                                        & .57      & .96                                          & .99  & 0.0 & 0.0 & 7.8  & $+0.830$ \\
    \midrule
    \multirow{6}{*}{\rotatebox[origin=c]{90}{ASAP}}  & GPT-5.4-nano    & .79                                        & .74      & .85                                          & .21  & 0.0 & 4.9 & 10.7 & $+0.677$ \\
                                                     & GPT-5.4-mini    & .73                                        & .62      & .84                                          & .03  & 0.0 & 0.0 & 13.8 & $+0.674$ \\
                                                     & Ministral-3-3B  & .78                                        & .65      & .94                                          & .02  & 0.0 & 0.4 & 7.1  & $+0.726$ \\
                                                     & Ministral-3-14B & .83                                        & .77      & .94                                          & .02  & 0.0 & 0.0 & 3.6  & $+0.690$ \\
                                                     & Gemma-3-4B      & .34                                        & .19      & .53                                          & 1.00 & 0.0 & 0.0 & 29.3 & $+0.668$ \\
                                                     & Gemma-3-12B     & .84                                        & .78      & .95                                          & .86  & 0.0 & 0.9 & 4.9  & $+0.735$ \\
    \bottomrule
  \end{tabular}

  \caption{%
    Inconsistency as a close-pair phenomenon does not track ranking accuracy.
    Consistency is how often the judge chooses the same text across the swapped pairs; IPP is inconsistent-pair primacy, with .5 indicating perfect balance. \textsc{close}/\textsc{far} are the same rate restricted to the bottom/top quartile of the gold gap $|g_a-g_b|$.  Triad-\% is the proportion of all triads. $\rho$ is the Spearman correlation of the LLM-derived BT skill against human gold.
  }
  \label{tab:posbias_close_far}
\end{table*}

\paragraph{Position Bias}
We evaluate GPT-5.4-nano and -mini, Ministral-3-3B and -14B, as well as Gemma-3-4B and -12B (see appendix~\ref{app:models}) on both the \textsc{Clear} and the \textsc{ASAP 2.0} corpus. Each pair is evaluated twice with reversed answer orderings. We measure position bias using two standard metrics: (1) the consistency, i.e., the frequency with which the same verdict is retained after position reversal; and (2) inconsistent-pair primacy (IPP), the proportion of inconsistent pairs in which the judge prefers the first-shown response. An IPP of .5 indicates that inconsistent verdicts split evenly between positions.

Dividing positional consistency into different $|\Delta|$-quartiles shows that position bias exists mostly on close pairs. Since a flipped verdict amounts to a tie under balanced aggregation, this reflects indecision on very close pairs ("too close to score").

Crucially, consistency, IPP, and rank correlation with gold do not track
each other (Tables~\ref{tab:posbias_close_far} and~\ref{tab:correlation}).
On \textsc{Clear}, the two GPTs flip at indistinguishable
rates\footnote{ Differences in consistency, IPP and intransitivity are tested with an exact McNemar test at the item level; differences in $\rho$ use a paired item bootstrap. Within each corpus, $p$-values are Holm-corrected; $\alpha = 0.05$.}, yet mini achieves significantly higher $\rho$. On \textsc{Asap} the
pattern reverses: nano is significantly more consistent while the two rank
the field equally well. Ministral behaves similarly on \textsc{Asap};
only on \textsc{Clear} do its bias and correlation align in the expected
direction. This is evidence that aggregate bias metrics are weak predictors
of ranking performance, not that bias is benign.

Gemma-3-4B is the limiting case: almost entirely indecisive on close pairs,
yet its far-pair judgments suffice to recover a ranking comparable to much
more consistent judges. Its ceiling is naturally capped --- it cannot
resolve the "middle pack" --- but the bias metric overstates the damage.
The mild top-quartile inconsistency present in every model points instead
to genuine conceptual misalignment: some pairs humans rank far apart are
treated by the model as close. Fixing that is the more promising lever for
correlation gains.

\paragraph{Non-Transitivity}
We analyze non-transitivity only on triads, distinguishing (1) cyclical
($A\prec B$, $B\prec C$, $C\prec A$), (2) mixed ($A\succ B$, $B\succ C$,
$C\sim A$), and (3) inequality ($A\sim B$, $B\sim C$, $A\not\sim C$). Given
the close-pair indecision documented above, inequality non-transitivity can
be substantively valid: if $|\Delta|$ between $A$ and $C$ is large enough
to score, the triad is consistent with the model's own behavior.

Across all judges, non-transitivity is dominated by the inequality type and
tracks inconsistency tightly (Pearson $r=.928$): it is largely a
re-expression of the indecision already captured by position bias. Within
each family the larger model generally produces fewer non-transitivities, so the
metric mainly tracks model size --- generally, but not reliably, a proxy
for judge quality. Despite far higher non-transitivity, Gemma-3-4B does
not correlate significantly worse with human gold than GPT-5.4-nano
(Table~\ref{tab:posbias_close_far}), confirming that intransitivity is a weak
model-selection signal.

\begin{table}[h]
  \centering
  \begin{tabular}{lcc}
    \toprule
    Proxy                     & $\rho_s$ vs ranking $\rho$ & $p_{\text{Holm}}$ \\
    \midrule
    Consistency, all          & $+0.559$                   & $0.198$           \\
    Consistency, close        & $+0.245$                   & $0.892$           \\
    \textbf{Consistency, far} & $+0.897$                   & $0.001^*$         \\
    Primacy (IPP)             & $+0.126$                   & $0.892$           \\
    Intransitivity            & $-0.608$                   & $0.174$           \\
    \bottomrule
  \end{tabular}
  \caption{Spearman correlation between inconsistencies and the model's ability
    to replicate human gold ranking, and the associated Holm-corrected
    significance. Two-sided permutation $p$ (10,000 reshuffles) over the 12
    judge-corpus points, Holm-corrected across the five proxies. With N=12 the
    test is well-powered for strong effects but has limited power against moderate
    effects. We note this as a limitation but emphasize the practical implication:
    a metric whose correlation with gold ranking is too weak to survive correction
    at this scale is too noisy to be used as a model-selection signal in practice.
  }  \label{tab:correlation}
\end{table}

\section{Conclusion}
Internal consistency and pairwise agreement with humans are useful judge properties, but the central question is whether a judge recovers the same latent ranking structure as humans. Therefore, the most direct evaluation criterion is the agreement between human and LLM rankings, and imperfect local consistency should not be treated as disqualifying. Aggregate inconsistency metrics such as position bias should be read in light of pair difficulty: a metric that weights all pairs equally rewards judges that excel on low-impact comparisons and penalizes those that excel on far-gap comparisons, which are more informative of judge quality.

This is not to say close pairs are irrelevant. Bias mitigation does help in making an already well-aligned judge more decisive on close pairs, which improves ranking granularity. Close pairs will matter more as judges improve and as tasks get harder (e.g. leaderboarding closely-performing state-of-the-art models). But since LLM judges still struggle on many tasks \cite{bavarescoLLMsInsteadHuman2025}, the far-pair regime is where gains are currently largest.

Two implications follow for benchmarks. For item design, judge accuracy scales with the rank gap between responses, so prompts that are difficult \textit{for the models being judged} are exactly those that yield pairs that are easy \textit{for an aligned judge}. This offers an explanation for the separability reported for benchmarks such as ArenaHard \cite{li2024crowdsourced}. However, item pools that elicit uniformly strong or uniformly weak responses both compress the gap, so item difficulty should be designed for response spread, not for higher difficulty as such. For gold standards, pair difficulty should be made observable. Items should be organized into rankings rather than isolated pairs wherever possible, and where that is infeasible, pair-difficulty-aware metrics should be reported. We do not recommend a specific metric, however, since the choice depends on how the human gold pairs were elicited and needs new behavioral data. For example, annotators could rate pair closeness directly, but this presupposes that humans can estimate closeness reliably; alternatively, annotator disagreement could proxy for difficulty, but how well disagreement tracks the rank gap has, to our knowledge, not been systematically studied. Instead of chasing ever-lower bias and ever-higher pairwise agreement, the next generation of judge evaluations should target the regimes where ranking signal actually lives.

\section*{Limitations}
Our rank-gap analysis requires a notion of latent strength on each corpus. \textsc{Clear} comes with BT-scores pre-computed. On \textsc{ASAP 2.0} we use pointwise human ratings as a strength proxy; this is a noisier estimator of $\theta^{\mathrm{human}}$ than pairwise-derived strengths and weakens within-bucket contrasts, so the reported pattern on \textsc{ASAP 2.0} is a lower bound. Specifically, real ties, though likely rare, cannot be distinguished from a quality gap that is below the rating resolution. We therefore also provide a reanalysis excluding gold ties in appendix~\ref{app:asap}.

The datasets we used both come from the educational domain, and so our results may not generalize to certain settings where LLM-as-a-judge is often used, such as QA and leaderboards. We made this choice because the two datasets provide text-level ground truth, which is needed to measure pair difficulty. For these other settings, to our knowledge, there do not exist datasets with item-level ground truth we could have evaluated on.

We restrict our attention to pairwise judging. The theoretical argument is specific to binary comparison, and listwise judging introduces additional position-bias structure \cite{shiJudgingJudgesSystematic2025} that the present framework does not address.

Finally, our experiments assume the goal of LLM-as-a-judge is to recover a human gold standard that is usually somewhat subjective.
Human raters are themselves imperfect estimators of any underlying quality, and what they rate may diverge from what we actually care about estimating --- on \textsc{Clear}, for instance, teacher ratings of text comprehensibility are not the same construct as comprehension measured directly from student readers. In settings with an objective gold and well-separated candidates, such as verifiable math or code tasks with skill-stratified models, the close-pair regime our analysis emphasizes becomes sparse and the practical importance of the misweighting effect diminishes, though the underlying theory remains unchanged.

\section*{Acknowledgements}
We thank Mayumi Ohta for identifying a flaw in our original account of why far pairs tend to matter more, which led to the revision in Section~3.

\section*{Use of AI Assistants}
AI writing assistance was used for sentence-level paraphrasing and polishing of author-written text; it was not used to generate research ideas, technical claims, related-work positioning, or citations. AI-assisted coding tools used during analysis script development are documented in the released code repository's README.

\bibliography{custom}

\begin{thebibliography}{19}
\providecommand{\natexlab}[1]{#1}

\bibitem[{Bavaresco et~al.(2025)Bavaresco, Bernardi, Bertolazzi, Elliott,
  Fern{\'a}ndez, Gatt, Ghaleb, Giulianelli, Hanna, Koller, Martins, Mondorf,
  Neplenbroek, Pezzelle, Plank, Schlangen, Suglia, Surikuchi, Takmaz, and
  Testoni}]{bavarescoLLMsInsteadHuman2025}
Anna Bavaresco, Raffaella Bernardi, Leonardo Bertolazzi, Desmond Elliott,
  Raquel Fern{\'a}ndez, Albert Gatt, Esam Ghaleb, Mario Giulianelli, Michael
  Hanna, Alexander Koller, Andre Martins, Philipp Mondorf, Vera Neplenbroek,
  Sandro Pezzelle, Barbara Plank, David Schlangen, Alessandro Suglia, Aditya~K
  Surikuchi, Ece Takmaz, and Alberto Testoni. 2025.
\newblock \href {https://doi.org/10.18653/v1/2025.acl-short.20} {{{LLMs}}
  instead of {{Human Judges}}? {{A Large Scale Empirical Study}} across 20
  {{NLP Evaluation Tasks}}}.
\newblock In \emph{Proceedings of the 63rd {{Annual Meeting}} of the
  {{Association}} for {{Computational Linguistics}} ({{Volume}} 2: {{Short
  Papers}})}, pages 238--255, Vienna, Austria.

\bibitem[{Bradley and Terry(1952)}]{bradleyRankAnalysisIncomplete1952}
Ralph~Allan Bradley and Milton~E. Terry. 1952.
\newblock \href {https://doi.org/10.2307/2334029} {Rank {{Analysis}} of
  {{Incomplete Block Designs}}: {{I}}. {{The Method}} of {{Paired
  Comparisons}}}.
\newblock \emph{Biometrika}, 39(3/4):324--345.

\bibitem[{Crossley et~al.(2023)Crossley, Heintz, Choi, Batchelor, Karimi, and
  Malatinszky}]{crossleyCommonLitEaseReadability}
Scott Crossley, Aron Heintz, Joon~Suh Choi, Jordan Batchelor, Mehrnoush Karimi,
  and Agnes Malatinszky. 2023.
\newblock A large-scaled corpus for assessing text readability.
\newblock \emph{Behavior Research Methods}, 55(2):491--507.

\bibitem[{Crossley et~al.(2025)Crossley, Baffour, Burleigh, and
  King}]{crossleyLargescaleCorpusAssessing2025a}
Scott~A. Crossley, Perpetual Baffour, L.~Burleigh, and Jules King. 2025.
\newblock \href {https://doi.org/10.1016/j.asw.2025.100954} {A {{Large-Scale
  Corpus}} for {{Assessing Source-Based Writing Quality}}: {{ASAP}} 2.0}.
\newblock \emph{Assessing Writing}, 65:100954.

\bibitem[{Daynauth et~al.(2025)Daynauth, Clarke, Flautner, Tang, and
  Mars}]{daynauthRankingUnraveledRecipes2025}
Roland Daynauth, Christopher Clarke, Krisztian Flautner, Lingjia Tang, and
  Jason Mars. 2025.
\newblock \href {https://doi.org/10.18653/v1/2025.acl-long.1265} {Ranking
  {{Unraveled}}: {{Recipes}} for {{LLM Rankings}} in {{Head-to-Head AI
  Combat}}}.
\newblock In \emph{Proceedings of the 63rd {{Annual Meeting}} of the
  {{Association}} for {{Computational Linguistics}} ({{Volume}} 1: {{Long
  Papers}})}, pages 26078--26091, Vienna, Austria.

\bibitem[{Fein et~al.(2026)Fein, Russo, Xiang, Jolly, Rafailov, and
  Haber}]{feinLitBenchBenchmarkDataset2026}
Daniel Fein, Sebastian Russo, Violet Xiang, Kabir Jolly, Rafael Rafailov, and
  Nick Haber. 2026.
\newblock \href {https://doi.org/10.18653/v1/2026.eacl-long.362} {{{LitBench}}:
  {{A Benchmark}} and {{Dataset}} for {{Reliable Evaluation}} of {{Creative
  Writing}}}.
\newblock In \emph{Proceedings of the 19th {{Conference}} of the {{European
  Chapter}} of the {{Association}} for {{Computational Linguistics}}
  ({{Volume}} 1: {{Long Papers}})}, pages 7740--7755, Rabat, Morocco.

\bibitem[{Feng et~al.(2025)Feng, Wang, Cheng, Wan, and
  Chen}]{fengSageScalableFramework2025}
Yuanning Feng, Sinan Wang, Zhengxiang Cheng, Yao Wan, and Dongping Chen. 2025.
\newblock \href {https://arxiv.org/abs/2512.16041} {Are {{We}} on the {{Right
  Way}} to {{Assessing LLM-as-a-Judge}}?}
\newblock \emph{Preprint}, arXiv:2512.16041.

\bibitem[{{{Gemma Team}} et~al.(2025){{Gemma Team}}, Kamath, Ferret, Pathak,
  Vieillard, Merhej, Perrin, Matejovicova, Ramé, Rivière, Rouillard, Mesnard,
  Cideron, bastien Grill, Ramos, Yvinec, Casbon, Pot, Penchev, Liu, Visin,
  Kenealy, Beyer, Zhai, Tsitsulin, Busa-Fekete, Feng, Sachdeva, Coleman, Gao,
  Mustafa, Barr, Parisotto, Tian, Eyal, Cherry, Peter, Sinopalnikov,
  Bhupatiraju, Agarwal, Kazemi, Malkin, Kumar, Vilar, Brusilovsky, Luo,
  Steiner, Friesen, Sharma, Sharma, Gilady, Goedeckemeyer, Saade, Feng,
  Kolesnikov, Bendebury, Abdagic, Vadi, György, Pinto, Das, Bapna, Miech,
  Yang, Paterson, Shenoy, Chakrabarti, Piot, Wu, Shahriari, Petrini, Chen, Lan,
  Choquette-Choo, Carey, Brick, Deutsch, Eisenbud, Cattle, Cheng, Paparas,
  Sreepathihalli, Reid, Tran, Zelle, Noland, Huizenga, Kharitonov, Liu,
  Amirkhanyan, Cameron, Hashemi, Klimczak-Plucińska, Singh, Mehta, Lehri,
  Hazimeh, Ballantyne, Szpektor, Nardini, Pouget-Abadie, Chan, Stanton,
  Wieting, Lai, Orbay, Fernandez, Newlan, yeong Ji, Singh, Black, Yu, Hui,
  Vodrahalli, Greff, Qiu, Valentine, Coelho, Ritter, Hoffman, Watson,
  Chaturvedi, Moynihan, Ma, Babar, Noy, Byrd, Roy, Momchev, Chauhan, Sachdeva,
  Bunyan, Botarda, Caron, Rubenstein, Culliton, Schmid, Sessa, Xu, Stanczyk,
  Tafti, Shivanna, Wu, Pan, Rokni, Willoughby, Vallu, Mullins, Jerome, Smoot,
  Girgin, Iqbal, Reddy, Sheth, Põder, Bhatnagar, Panyam, Eiger, Zhang, Liu,
  Yacovone, Liechty, Kalra, Evci, Misra, Roseberry, Feinberg, Kolesnikov, Han,
  Kwon, Chen, Chow, Zhu, Wei, Egyed, Cotruta, Giang, Kirk, Rao, Black, Babar,
  Lo, Moreira, Martins, Sanseviero, Gonzalez, Gleicher, Warkentin, Mirrokni,
  Senter, Collins, Barral, Ghahramani, Hadsell, Matias, Sculley, Petrov,
  Fiedel, Shazeer, Vinyals, Dean, Hassabis, Kavukcuoglu, Farabet, Buchatskaya,
  Alayrac, Anil, Dmitry, Lepikhin, Borgeaud, Bachem, Joulin, Andreev, Hardin,
  Dadashi, and Hussenot}]{gemmateam2025gemma3technicalreport}
{{Gemma Team}}, Aishwarya Kamath, Johan Ferret, Shreya Pathak, Nino Vieillard,
  Ramona Merhej, Sarah Perrin, Tatiana Matejovicova, Alexandre Ramé, Morgane
  Rivière, Louis Rouillard, Thomas Mesnard, Geoffrey Cideron, Jean bastien
  Grill, Sabela Ramos, Edouard Yvinec, Michelle Casbon, Etienne Pot, Ivo
  Penchev, and 197 others. 2025.
\newblock \href {https://arxiv.org/abs/2503.19786} {Gemma 3 {{Technical
  Report}}}.
\newblock \emph{Preprint}, arXiv:2503.19786.

\bibitem[{Gu et~al.(2026)Gu, Jiang, Shi, Tan, Zhai, Xu, Li, Shen, Ma, Liu,
  Wang, Zhang, Lin, Zhang, Ni, Gao, Wang, and Guo}]{guSurveyLLMasaJudge2025a}
Jiawei Gu, Xuhui Jiang, Zhichao Shi, Hexiang Tan, Xuehao Zhai, Chengjin Xu, Wei
  Li, Yinghan Shen, Shengjie Ma, Honghao Liu, Saizhuo Wang, Kun Zhang, Zhouchi
  Lin, Bowen Zhang, Lionel Ni, Wen Gao, Yuanzhuo Wang, and Jian Guo. 2026.
\newblock \href {https://doi.org/10.1016/j.xinn.2025.101253} {A survey on
  {{LLM}}-as-a-judge}.
\newblock \emph{The Innovation}, 7(6):101253.

\bibitem[{Li et~al.(2024)Li, Sun, Yuan, Fan, Zhao, and
  Liu}]{liGenerativeJudgeEvaluating2023}
Junlong Li, Shichao Sun, Weizhe Yuan, Run-Ze Fan, Hai Zhao, and Pengfei Liu.
  2024.
\newblock \href
  {https://proceedings.iclr.cc/paper_files/paper/2024/file/747dc7c6566c74eb9a663bcd8d057c78-Paper-Conference.pdf}
  {{{Generative Judge}} for {{Evaluating Alignment}}}.
\newblock In \emph{International Conference on Learning Representations},
  volume 2024, pages 27547--27574.

\bibitem[{Li et~al.(2025)Li, Chiang, Frick, Dunlap, Wu, Zhu, Gonzalez, and
  Stoica}]{li2024crowdsourced}
Tianle Li, Wei-Lin Chiang, Evan Frick, Lisa Dunlap, Tianhao Wu, Banghua Zhu,
  Joseph~E. Gonzalez, and Ion Stoica. 2025.
\newblock \href {https://openreview.net/forum?id=KfTf9vFvSn} {From crowdsourced
  data to high-quality benchmarks: Arena-hard and benchbuilder pipeline}.
\newblock In \emph{Proceedings of the 42nd {{International Conference}} on
  {{Machine Learning}}}, volume 267 of \emph{Proceedings of Machine Learning
  Research}, pages 34209--34231, Vancouver, Canada. PMLR.

\bibitem[{Liu et~al.(2026)Liu, Khandelwal, Subramanian, Jouault, Rastogi,
  Sadé, Jeffares, Jiang, Cahill, Gavaudan, Sablayrolles, Héliou, You,
  Ehrenberg, Lo, Eliseev, Calvi, Sooriyarachchi, Bout, Rozière, Monicault,
  Lanfranchi, Barreau, Courtot, Grattarola, Dabert, de~las Casas, Chane-Sane,
  Ahmed, Berrada, Ecrepont, Guinet, Novikov, Kunsch, Lample, Martin, Gupta,
  Ludziejewski, Rute, Studnia, Amar, Delas, Roberts, Yadav, Chandu, Jain,
  Aitchison, Fainsin, Blier, Zhao, Martin, Saulnier, Gao, Buyl, Jennings,
  Pellat, Prins, Poirée, Guillaumin, Dinot, Futeral, Darrin, Augustin,
  Chiquier, Schimpf, Grinsztajn, Gupta, Raghuraman, Bousquet, Duchenne, Wang,
  von Platen, Jacob, Wambergue, Kurylowicz, Muddireddy, Chagniot, Stock,
  Agrawal, Torroba, Sauvestre, Soletskyi, Menneer, Vaze, Barry, Gandhi,
  Waghjale, Gandhi, Ghosh, Mishra, Aithal, Antoniak, Scao, Cachet, Sorg,
  Lavril, Saada, Chabal, Foubert, Robert, Wang, Lawson, Bewley, Bewley,
  Edwards, Jamil, Tomasini, Nemychnikova, Phung, Maladière, Richard, Bouaziz,
  Li, Marshall, Li, Yang, Ouahidi, Wang, Tang, and Ramzi}]{liu2026ministral3}
Alexander~H. Liu, Kartik Khandelwal, Sandeep Subramanian, Victor Jouault,
  Abhinav Rastogi, Adrien Sadé, Alan Jeffares, Albert Jiang, Alexandre Cahill,
  Alexandre Gavaudan, Alexandre Sablayrolles, Amélie Héliou, Amos You, Andy
  Ehrenberg, Andy Lo, Anton Eliseev, Antonia Calvi, Avinash Sooriyarachchi,
  Baptiste Bout, and 101 others. 2026.
\newblock \href {https://arxiv.org/abs/2601.08584} {Ministral 3}.
\newblock \emph{Preprint}, arXiv:2601.08584.

\bibitem[{{OpenAI}(2026)}]{IntroducingGPT54Mini2026}
{OpenAI}. 2026.
\newblock Introducing {{GPT-5}}.4 mini and nano.
\newblock {{https://openai.com/index/introducing-gpt-5-4-mini-and-nano/}}.

\bibitem[{Shi et~al.(2025)Shi, Ma, Liang, Diao, Ma, and
  Vosoughi}]{shiJudgingJudgesSystematic2025}
Lin Shi, Chiyu Ma, Wenhua Liang, Xingjian Diao, Weicheng Ma, and Soroush
  Vosoughi. 2025.
\newblock \href {https://doi.org/10.18653/v1/2025.ijcnlp-long.18} {{{Judging}}
  the {{Judges}}: {{A Systematic Study}} of {{Position Bias}} in
  {LLM}-as-a-{{Judge}}}.
\newblock In \emph{Proceedings of the 14th International Joint Conference on
  Natural Language Processing and the 4th Conference of the Asia-Pacific
  Chapter of the Association for Computational Linguistics}, pages 292--314,
  Mumbai, India. The Asian Federation of Natural Language Processing and The
  Association for Computational Linguistics.

\bibitem[{Thakur et~al.(2025)Thakur, Choudhary, Ramayapally, Vaidyanathan, and
  Hupkes}]{thakurJudgingJudgesEvaluating2025}
Aman~Singh Thakur, Kartik Choudhary, Venkat~Srinik Ramayapally, Sankaran
  Vaidyanathan, and Dieuwke Hupkes. 2025.
\newblock Judging the {{Judges}}: {{Evaluating Alignment}} and
  {{Vulnerabilities}} in {{LLMs-as-Judges}}.
\newblock In \emph{Proceedings of the {{Fourth Workshop}} on {{Generation}},
  {{Evaluation}} and {{Metrics}} ({{GEM}}{$^2$})}, pages 404--430, Vienna,
  Austria.

\bibitem[{Wang et~al.(2024)Wang, Li, Chen, Cai, Zhu, Lin, Cao, Kong, Liu, Liu,
  and Sui}]{wangLargeLanguageModels2024}
Peiyi Wang, Lei Li, Liang Chen, Zefan Cai, Dawei Zhu, Binghuai Lin, Yunbo Cao,
  Lingpeng Kong, Qi~Liu, Tianyu Liu, and Zhifang Sui. 2024.
\newblock \href {https://doi.org/10.18653/v1/2024.acl-long.511} {Large
  {{Language Models}} are not {{Fair Evaluators}}}.
\newblock In \emph{Proceedings of the 62nd {{Annual Meeting}} of the
  {{Association}} for {{Computational Linguistics}} ({{Volume}} 1: {{Long
  Papers}})}, pages 9440--9450, Bangkok, Thailand.

\bibitem[{Wang et~al.(2026)Wang, Song, Zhu, Zhang, Yu, Chen, Song, Wang, Wu,
  Dai, Zhang, Wang, Ye, and
  Zhang}]{wangTrustJudgeInconsistenciesLLMasaJudge2025}
Yidong Wang, Yunze Song, Tingyuan Zhu, Xuanwang Zhang, Zhuohao Yu, Hao Chen,
  Chiyu Song, Qiufeng Wang, Zhen Wu, Xinyu Dai, Yue Zhang, Cunxiang Wang, Wei
  Ye, and Shikun Zhang. 2026.
\newblock \href {https://openreview.net/forum?id=4uPyOCeN6U} {{TrustJudge}:
  Inconsistencies of {LLM}-as-a-judge and how to alleviate them}.
\newblock In \emph{The Fourteenth International Conference on Learning
  Representations (ICLR)}.

\bibitem[{Xu et~al.(2025)Xu, Ruis, Rockt{\"a}schel, and
  Kirk}]{xuInvestigatingNonTransitivityLLMasaJudge2025}
Yi~Xu, Laura Ruis, Tim Rockt{\"a}schel, and Robert Kirk. 2025.
\newblock \href {https://proceedings.mlr.press/v267/xu25w.html} {Investigating
  {{Non-Transitivity}} in {{LLM-as-a-Judge}}}.
\newblock In \emph{Proceedings of the 42nd {{International Conference}} on
  {{Machine Learning}}}, volume 267 of \emph{Proceedings of Machine Learning
  Research}, pages 69583--69612, Vancouver, Canada. PMLR.

\bibitem[{Zheng et~al.(2023)Zheng, Chiang, Sheng, Zhuang, Wu, Zhuang, Lin, Li,
  Li, Xing, Zhang, Gonzalez, and Stoica}]{zhengJudgingLLMasaJudgeMTBench2023a}
Lianmin Zheng, Wei-Lin Chiang, Ying Sheng, Siyuan Zhuang, Zhanghao Wu, Yonghao
  Zhuang, Zi~Lin, Zhuohan Li, Dacheng Li, Eric Xing, Hao Zhang, Joseph
  Gonzalez, and Ion Stoica. 2023.
\newblock \href {https://doi.org/10.52202/075280-2020} {Judging
  {{LLM}}-as-a-{{Judge}} with {{MT}}-{{Bench}} and {{Chatbot Arena}}}.
\newblock In \emph{Advances in Neural Information Processing Systems},
  volume~36, pages 46595--46623. Curran Associates, Inc.

\end{thebibliography}

\appendix

\section{Models}
\label{app:models}

We evaluate six judges spanning three model families and two size
tiers per family. Table~\ref{tab:models} summarizes parameter counts,
licenses, and access modalities. All model use is consistent with the
intended research and evaluation use stated by each provider. We focus on smaller models (3B–14B parameters) because the variance-limited regime our analysis targets is most pronounced at this scale; larger models exhibit ceiling effects on the corpora we use.

\begin{table}[h]
  \centering
  \small
  \setlength{\tabcolsep}{4pt}
  \begin{tabular}{lll}
    \toprule
    Model & Params & Access      \\
    \midrule
    \makecell[l]{GPT-5.4 nano    \\ \cite{IntroducingGPT54Mini2026}}    & undisclosed & API \\
    \makecell[l]{GPT-5.4 mini    \\ \cite{IntroducingGPT54Mini2026}}    & undisclosed & API \\
    \makecell[l]{Ministral 3 3B  \\ \cite{liu2026ministral3}}         & 3B          & Open weights \\
    \makecell[l]{Ministral 3 14B \\ \cite{liu2026ministral3}}        & 14B         & Open weights \\
    \makecell[l]{Gemma 3 4B      \\ \cite{gemmateam2025gemma3technicalreport}}      & 4B          & Open weights \\
    \makecell[l]{Gemma 3 12B     \\ \cite{gemmateam2025gemma3technicalreport}}     & 12B         & Open weights \\
    \bottomrule
  \end{tabular}
  \caption{Judge models.}
  \label{tab:models}
\end{table}

Note that this generation of OpenAI models does not support fully deterministic output (which we used for the open weight models), which may inflate the reported inconsistency numbers and prevent exact replication. Nevertheless, these models are state-of-the-art (e.g., mini's top performance on \textsc{Clear}) and widely used in production, so we felt it was important to include them in order to situate API-based models within our broader thesis.

All open-weights inference was run locally on Apple Silicon using
llama.cpp with Q8\_0 quantization, which is deterministic under greedy
decoding and introduces minimal precision loss relative to FP16.
Total wall-clock across the four open-weights judges and two corpora
was approximately 20 hours.

\section{Prompts}

\subsection{\textsc{CLEAR}: readability ease}
Note: Exact instructions used for the human rating.

\begin{promptbox}{System prompt (\textsc{CLEAR})}
\raggedright
You are a school teacher.
\label{box:prompt-clear-sys}
\end{promptbox}

\begin{promptbox}{Pairwise judging prompt (\textsc{CLEAR})}
\raggedright
Which text is easier for students to understand?

\medskip

Text A: \\
\{text\_a\}

\medskip

----- \\
Text B: \\
\{text\_b\}

\medskip

-----
\label{box:prompt-clear-user}
\end{promptbox}

\subsection{\textsc{ASAP 2.0}: holistic essay quality}
Reformulates the pointwise criteria of the human raters into pairwise criteria.

\begin{promptbox}{System prompt (\textsc{ASAP 2.0})}
\raggedright
You are an expert essay grader comparing the overall (holistic) quality of two source-based, argumentative student essays written in response to the same writing prompt. Pick the higher-quality essay using the criteria below.
\label{box:prompt-asap-2-0-sys}
\end{promptbox}

\begin{promptbox}{Pairwise judging prompt (\textsc{ASAP 2.0})}
\raggedright
You are judging the overall (holistic) quality of a source-based, argumentative essay written by a school student in response to the writing prompt shown below. Quality is holistic: weigh everything that goes into a strong essay together, not any single feature in isolation.

\medskip

What contributes to holistic quality: \\
- whether the essay develops a clear, well-supported point of view on the issue, with genuine critical thinking (vs. a vague, simplistic, or missing position) \\
- whether it uses appropriate, accurate examples, reasons, and evidence drawn from the source text(s) (vs. sparse, misused, or merely personal-opinion support) \\
- whether it is well organized and focused, with coherence and smooth progression of ideas (vs. disjointed or unfocused) \\
- command of language: varied, accurate, apt vocabulary and meaningful sentence variety (vs. weak word choice or monotonous/broken structure) \\
- control of grammar, usage, and mechanics (vs. errors frequent or serious enough to obscure meaning)

\medskip

---

\medskip

The students first read the source article below, then wrote their essay in response to the writing prompt that follows. Use it to judge whether the essay's evidence is accurate and genuinely drawn from the source (vs. vague, misremembered, or fabricated).

\medskip

Source article — {\textquotedbl}Making Mona Lisa Smile{\textquotedbl} by Nick D'Alto:

\medskip

[full article text omitted here, see \cite{crossleyLargescaleCorpusAssessing2025a}]

\medskip

---

\medskip

Writing prompt the students responded to: \\
\{assignment\}

\medskip

---

\medskip

Essay A: \\
\{text\_a\}

\medskip

---

\medskip

Essay B: \\
\{text\_b\}

\medskip

---

\medskip

Which essay is higher quality overall?
\label{box:prompt-asap-2-0-user}
\end{promptbox}

\section{Simulation Under Different Settings}
\label{app:sim}

The simulation in section~\ref{sim} makes several assumptions (BT-behaving human, normal strength distribution, LLM-human pairwise agreement of only .6, judges that do not tie), but our code also allows for other simulation settings, which was run as a robustness check.

\begin{itemize}
  \item \textbf{Deterministic human judge.} The human picks the higher-gold item with fixed probability $p$ rather than according to BT probabilities, deterministically at $p=1$. Every judge disagreement is now an outright error with respect to gold, so far-pair disagreement is even more damaging for ranking recovery. Close-pair disagreements are errors in the same sense, but cost almost no ranking quality.
  \item \textbf{Uniform strength distribution.} A uniform distribution yields fewer close pairs and more far pairs relative to the normal baseline, so far-pair performance carries an even larger share of the ranking signal, and ranking recovery separates more strongly.
  \item \textbf{Higher pairwise agreement.} As agreement rises, all simulated judges approach the rank-correlation ceiling; the ranking correlation differences between them shrink, and pair difficulty matters less.
\end{itemize}

\begin{table*}[t]
  \centering
  \begin{tabular}{lccccc}
    \toprule
                    & \multicolumn{3}{c}{Consistency ($1-$flip)} & \multicolumn{2}{c}{$\rho$ vs gold}                                        \\
    \cmidrule(lr){2-4}\cmidrule(lr){5-6}
    Judge           & close                                      & close                              & far            & all      & tied     \\
                    & (gap $=0$)                                 & (gap $=1$)                         & (gap $\geq 2$) & verdicts & excluded \\
                    & 28.2\%                                     & 44.9\%                             & 26.8\%         &          &          \\
    \midrule
    GPT-5.4-nano    & .74                                        & .78                                & .85            & $+0.677$ & $+0.735$ \\
    GPT-5.4-mini    & .62                                        & .73                                & .84            & $+0.674$ & $+0.726$ \\
    Ministral-3-3B  & .65                                        & .76                                & .94            & $+0.726$ & $+0.786$ \\
    Ministral-3-14B & .77                                        & .80                                & .94            & $+0.690$ & $+0.753$ \\
    Gemma-3-4B      & .19                                        & .32                                & .53            & $+0.668$ & $+0.678$ \\
    Gemma-3-12B     & .78                                        & .81                                & .95            & $+0.735$ & $+0.797$ \\
    \bottomrule
  \end{tabular}

  \caption{\textsc{ASAP 2.0} robustness check with gold-tied pairs removed. Consistency is reported for the three gold-gap buckets.}  \label{tab:asap-noties}
\end{table*}

\begin{figure}[htbp]
  \centering
  \includegraphics[width=1\linewidth]{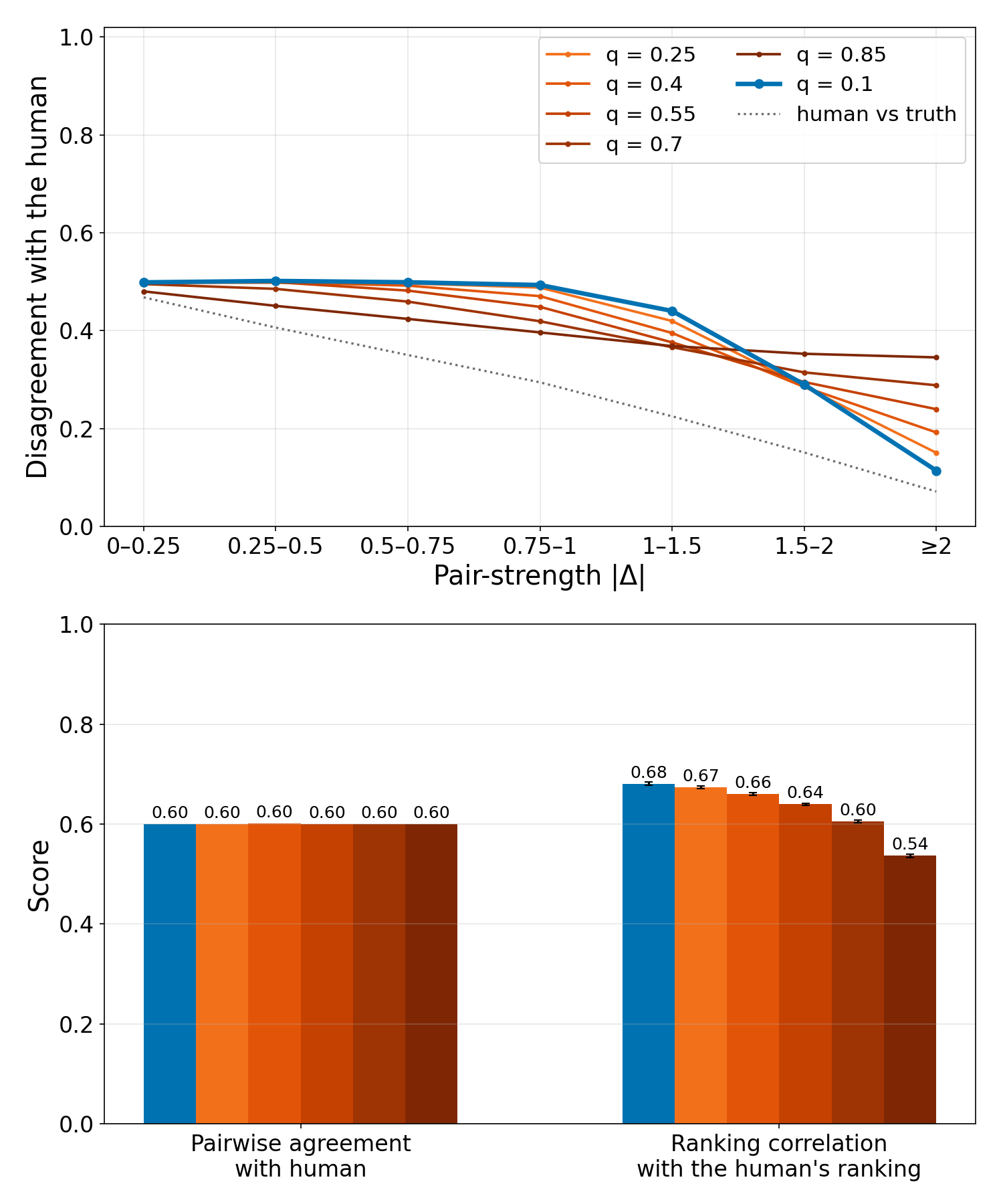}
  \caption{Simulation results for BT Judges with identical pairwise agreement that coinflip below a certain pair gap. Top shows how disagreement between judges and humans is differently distributed, while overall agreement is exactly equal (bottom left). Despite this, ranking correlation with the human gold varies substantially depending on where disagreement occurs (bottom right).}
  \label{fig:sim_tied}
\end{figure}

For a \textbf{judge that coinflips on close pairs instead of actively disagreeing}, the judges with identical agreement still recover the ranking differently, though the difference is lower. Naturally, fewer of the judges' disagreements can now come from close pairs, so more disagreements have to fall on far pairs even for the far-pair-aligned judges. We report this variant separately (Figure~\ref{fig:sim_tied}) since it changes both the agreement distribution as well as the ranking recovery.

\section{ASAP Without Gold-Tied Pairs}
\label{app:asap}

Due to integer pointwise scoring, ASAP~2.0 gold gaps have a low resolution, and ties are common. But a tie indicates a quality difference below rating resolution, not necessarily equality. The items may still have a quality difference that is imperceptible by the gold standard. But judge inconsistency is still measurable, unless we assume exact ties. We therefore report the numbers including ties in Table~\ref{tab:posbias_close_far}. But to confirm that no conclusion depends on these pairs, we repeat the ASAP analyses with them excluded (Table~\ref{tab:asap-noties}) as a robustness check.

The close bucket then becomes gap $=1$; the far bucket is unchanged (gap $\geq 2$). Consistency remains close $<$ far for all six judges. Ranking recovery does not degrade when gold-tied pairs are dropped from the BT input: $\rho$ against the human gold rises for every judge, by $+0.01$ to $+0.06$, likely because we are removing noise. The Table~\ref{tab:correlation} result is likewise unaffected, with Spearman(far-consistency, correlation with human gold) $=.897$ under the new binning, identical to the reported value.

\end{document}